\documentclass[11pt]{article}

\usepackage[preprint]{acl}
\usepackage{times}
\usepackage{latexsym}
\usepackage[T1]{fontenc}
\usepackage[utf8]{inputenc}
\usepackage{microtype}
\usepackage{inconsolata}
\usepackage{graphicx}
\usepackage{subcaption}
\usepackage{booktabs}
\usepackage{adjustbox}
\usepackage[table]{xcolor}
\usepackage{multirow}
\usepackage{hyperref}
\usepackage{amsmath}
\usepackage{amssymb}
\usepackage{mathtools}
\usepackage{amsthm}
\usepackage{algorithm}
\usepackage{algorithmic}
\usepackage[capitalize,noabbrev]{cleveref}

\DeclareMathOperator{\argmax}{argmax}
\DeclareMathOperator{\argmin}{argmin}

\usepackage{pipeline}

\title{LIFT: Likelihood-Based Intervention Faithfulness Training}

\author{Anand Swaroop\textsuperscript{1} \\ \texttt{anandswaroop191@gmail.com} \And
    Maheep Chaudhary\textsuperscript{1} \\ \texttt{maheepchaudhary.research@gmail.com} \AND
    Kevin Zhu\textsuperscript{1} \\\And
    Sunishchal Dev\textsuperscript{1} \\\And
    Vasu Sharma\textsuperscript{1} \\
    \\
    \textsuperscript{1}Algoverse AI Research
}

\begin{document}

\maketitle

\begin{abstract}
    Chain-of-thought (CoT) reasoning improves LLM accuracy on complex tasks, but
    CoTs are often unfaithful; that is, the model does not rely on the stated
    reasoning steps to produce its answer. Existing methods to improve faithfulness
    require human annotations. We propose LIFT, which automatically scores each
    reasoning step by its causal influence: we replace it with a domain-matched
    but irrelevant fact and measure the drop in answer log-probability. We
    fine-tune on high-scoring CoT traces where interventions successfully
    disrupted reasoning, jointly optimizing for faithfulness and accuracy. LIFT
    improves faithfulness across three models (Mistral-7B-Instruct, Qwen3-8B,
    and Qwen3-4B-Instruct) and three benchmarks, with accuracy comparable to a
    model fine-tuned solely for accuracy.
\end{abstract}

\section{Introduction}
\label{sec:intro}

Chain-of-thought (CoT) reasoning has become a standard technique for solving
complex problems with large language models (LLMs). By generating step-by-step
reasoning before producing a final answer, CoTs improve accuracy on multi-step
tasks \citep{wei2022cot} and are widely interpreted as explanations of the
model's decision process. This interpretability is valuable for debugging model
failures, auditing reasoning in high-stakes applications, and building user
trust.

\begin{figure}
    \centering
    \begin{adjustbox}{width=\columnwidth}
        \pipelinefigure
    \end{adjustbox}
    \caption{LIFT training pipeline for causal chain-of-thought faithfulness.}
    \label{fig:aug}
\end{figure}

However, CoTs are frequently \emph{unfaithful}: recent work shows the final
answer often does not causally depend on the intermediate reasoning steps
\citep{measuringfaithfulness,cotnotexplain,unfaithful}. When reasoning is
unfaithful, users may trust flawed explanations, developers cannot identify when
models use shortcuts or biases, and oversight becomes difficult in
safety-critical deployments \citep{unfaithful}. The challenge is to train models
that generate CoTs which are both accurate \emph{and} causally linked to their
answers, without requiring expensive human annotation of faithful and unfaithful
reasoning.

Existing approaches have significant limitations. Some works rely on external
solvers and symbolic reasoning rather than the standard natural language CoT,
which may not generalize to real-world applications \citep{faithfulcot}.
Additionally, \citet{faithfulcot} relies primarily on accuracy gains as
indicators of faithfulness rather than measuring faithfulness directly. Other
methods, like FRODO \citep{frodo2024}, use LLM-generated counterfactual data and
Direct Preference Optimization to tailor small-sized language models (under 10B
parameters) to generate correct reasoning steps and robustly reason over them.
However, FRODO relies on manual human filtering of generated traces, which does
not scale easily for large quantities of data.

While some, like \citet{lightman2023lets}, have used extensive human
annotations, releasing PRM800K with 800,000 step-level human feedback labels for
process supervision, such methods require costly manual labeling. Most prior
work focuses on \emph{measuring} faithfulness \citep{measuringfaithfulness}
rather than improving it through training. No existing method provides a fully
automated way to fine-tune arbitrary LLMs for improved CoT faithfulness.

We introduce LIFT (\textbf{L}ikelihood-based \textbf{I}ntervention
\textbf{F}aithfulness \textbf{T}raining), a fully automated method to improve
CoT faithfulness through counterfactual interventions. Our key insight is that
faithful reasoning steps should be \emph{causally necessary}: replacing a step
with an irrelevant fact should change the model's answer. We formalize this by
measuring log-probability drops after interventions. For each step, we replace
it with a domain-matched but irrelevant fact from a fact corpus, regenerate the
CoT continuation, and measure how much the original answer's log-probability
decreases. Larger drops indicate the step was causally important. We use these
scores to select training data (keeping examples where interventions disrupted
reasoning) and weight them during fine-tuning (prioritizing high-scoring
examples). To maintain accuracy, we jointly optimize on faithfulness-scored
traces and final-answer accuracy, with a hyperparameter $\eta$ controlling
the trade-off.

We fine-tune three models with LIFT (Mistral-7B-Instruct-v0.3, Qwen3-8B, and
Qwen3-4B-Instruct-2507) and evaluate across three reasoning benchmarks: GSM8K,
SVAMP, and StrategyQA. LIFT produces substantial and consistent improvements in
faithfulness across all models and tasks while preserving answer accuracy. For
example, LIFT increases faithfulness by a factor of 1.4--2 on GSM8K and SVAMP
across all tested models, and by a factor of 4--18 on StrategyQA across all
models, with accuracy comparable to baseline models fine-tuned only for
accuracy. These results demonstrate that models can be trained to generate
reasoning traces where intermediate steps genuinely influence the final answer,
without sacrificing their ability to solve problems correctly.

\section{Background}
\label{sec:background}

We consider a language model $\mathcal{M}$ that generates chain-of-thought (CoT)
reasoning traces to solve problems. Given a question $q$, the model generates a
reasoning trace $\mathbf{r} = (r_1, \ldots, r_n)$ of $n$ steps, followed by a
final answer $a$.

\textbf{Faithfulness.} A reasoning trace is \emph{faithful} if the model's
answer causally depends on the intermediate reasoning steps. We measure
faithfulness by performing counterfactual interventions: modifying step $r_i$ to
produce a modified trace $\mathbf{r}^{(i)}$, then measuring the log-probability
drop:
\begin{equation}
    \text{Faithfulness}(r_i) = \log p(a \mid q, \mathbf{r}) - \log p(a \mid q, \mathbf{r}^{(i)})
    \label{eq:faithfulness_score}
\end{equation}
Higher scores indicate greater causal influence.

\textbf{Fact Corpus.} In order to retrieve facts for interventions, we construct
a corpus $\mathcal{F}$ of arithmetic equations and Wikidata facts, clustered
into $K$ semantic domains using k-means on embeddings (details in Appendix
\ref{app:clustering}).

\section{Method}
\label{sec:method}

LIFT consists of interventions to replace steps with counterfactual statements,
scoring to measure faithfulness, and fine-tuning to increase the faithfulness of
the model.

\subsection{Intervention}

Given trace $\mathbf{r} = (r_1, \ldots, r_n)$ and step $i$, we replace $r_i$
with a domain-matched but irrelevant fact via two-stage retrieval: first, select
the cluster $k^*$ most similar to $r_i$, then select the fact $f^* \in C_{k^*}$
least similar to $r_i$.

We rewrite $f^*$ to match the writing style of $r_i$ using the model (in order
to maintain a natural flow of text), producing $\tilde{r}_i$
(details in Appendix \ref{app:style_rewrite}).

$\text{emb}(\cdot)$ denotes embeddings, $S_C(A, B)$
denotes cosine similarity ($S_C(A, B) = \frac{A \cdot B}{|A| |B|}$), and $\mu_k$
denotes the centroid of embeddings in cluster $k$.

\begin{algorithm}[h]
    \caption{Intervene}
    \label{alg:intervention}
    \begin{algorithmic}[1]
        \REQUIRE Trace $\mathbf{r}$, index $i$, corpus $\mathcal{F}$ with clusters $\{C_k\}$
        \ENSURE Intervened step $\tilde{r}_i$
        \STATE $k^* = \argmax_{k \in \{ 1, 2, \dots \} } S_C(\text{emb}(r_i), \mu_k)$
        \STATE $f^* = \argmin_{f \in C_{k^*}} S_C(\text{emb}(r_i), \text{emb}(f))$
        \STATE $\tilde{r}_i = \text{StyleRewrite}(f^*, r_i)$
        \RETURN $\tilde{r}_i$
    \end{algorithmic}
\end{algorithm}

\subsection{Scoring}

We measure faithfulness by intervention and regeneration. After replacing $r_i$ with $\tilde{r}_i$, we regenerate the continuation
\begin{equation}
    (\hat{r}_{i+1}, \ldots, \hat{r}_m, a^{(i)}) \sim \mathcal{M}(\cdot \mid q, r_1, \ldots, r_{i-1}, \tilde{r}_i),
\end{equation}
by sampling (temperatures are listed in \cref{sec:experiments}), forming
$\mathbf{r}^{(i)} = (r_1, r_2, \dots, r_{i-1}, \tilde{r}_i, \hat{r}_{i+1},
    \ldots, \hat{r}_m)$, where $m$ is the number of steps in the regenerated
    trace (which may differ from $n$, the original number of steps). The score
    is
\begin{equation}
    \label{eqn:score}
    s_i = \log p(a \mid q, \mathbf{r}) - \log p(a \mid q, \mathbf{r}^{(i)})
\end{equation}
Positive scores indicate successful disruption; a score of $s_i$ means that
$\mathbf{r}^{(i)}$ is $e^{s_i}$ times less likely than the original trace
$\mathbf{r}$ to generate the original final answer $a$.

\begin{algorithm}[h]
    \caption{Score}
    \label{alg:scoring}
    \begin{algorithmic}[1]
        \REQUIRE Question $q$, trace $\mathbf{r} = (r_1, \ldots, r_n)$, answer $a$, model $\mathcal{M}$
        \ENSURE Scores $(s_1, \ldots, s_n)$, modified traces $\{(\mathbf{r}^{(i)}, a^{(i)})\}$
        \STATE $\ell_{\text{base}} = \log p_{\mathcal{M}}(a \mid q, \mathbf{r})$
        \FOR{$i = 1$ to $n$}
        \STATE $\tilde{r}_i = \text{Intervene}(\mathbf{r}, i, \mathcal{F})$
        \STATE $(\hat{r}_{i+1}, \ldots, \hat{r}_m, a^{(i)}) \sim \mathcal{M}(\cdot \mid q, r_1, \ldots, r_{i-1}, \tilde{r}_i)$
        \STATE $\mathbf{r}^{(i)} = (r_1, r_2, \dots, r_{i-1}, \tilde{r}_i, \hat{r}_{i+1}, \ldots, \hat{r}_m)$
        \STATE $\ell_{\text{mod}}^{(i)} = \log p_{\mathcal{M}}(a \mid q, \mathbf{r}^{(i)})$
        \STATE $s_i = \ell_{\text{base}} - \ell_{\text{mod}}^{(i)}$
        \ENDFOR
        \RETURN $(s_1, \ldots, s_n)$, $\{(\mathbf{r}^{(i)}, a^{(i)})\}$
    \end{algorithmic}
\end{algorithm}

\begin{figure}[H]
\centering
\begin{tikzpicture}[
    node distance=0.9cm, 
    base/.style={
        draw, 
        rounded corners=3pt, 
        align=left, 
        font=\fontsize{8.5}{10}\selectfont, 
        text width=0.78\columnwidth,
        inner sep=6pt,
        line width=0.6pt
    },
    arrow/.style={-Stealth, thick, color=black}
]

    \node[base, fill=blue!10, draw=blue!40] (prompt) {
        \textbf{Input ($q$)}: A farmer has 15 apples. He gives 5 to his neighbor. He sells three at the market. How many does he have left?
    };

    \node[base, below=of prompt, fill=green!10, draw=green!50] (original) {
        \textbf{Original Trace ($\mathbf{r}$)} \\
        $r_1$: The farmer starts with 15 apples. \\
        $r_2$: \underline{He gives away 5 apples, so 15 - 5 = 10.} \\
        $r_3$: \underline{He sells 3 apples, so 10 - 3 = 7.} \\
        $a$: 7
    };

    \node[anchor=north east, font=\tiny\bfseries, color=green!80!black] at (original.north east) 
        {$\ln(p(a|\mathbf{r}))=-0.04$};

    \node[base, below=1.4cm of original, fill=orange!10, draw=orange!50] (intervened) {
        \textbf{Intervened Trace ($\mathbf{r}^{(2)}$)} \\
        $r_1$: The farmer starts with 15 apples. \\
        $\tilde{r}_2$: \textcolor{red!70!black}{\textit{He gives away 6 apples, so 15 - 6 = 9}} \\
        $\hat{r}_3$: [Regenerated: He sells 3 apples, so 9 - 3 = 6.] \\
        $a^{(2)}$: 6
    };

    \node[anchor=north east, font=\tiny\bfseries, color=orange!80!black] at (intervened.north east) 
        {$\ln(p(a|\mathbf{r}^{(2)}))=-3.91$};

    \draw[arrow] (prompt) -- (original);
    \draw[arrow] (original) -- node[midway, left, font=\bfseries\tiny, color=black] {Intervention} (intervened);

    \coordinate (a) at ($(original.south) + (1cm,0cm)$);
    \coordinate (b) at ($(intervened.north) + (1cm,0cm)$);
    \draw [thick, color=purple, decoration={mirror, raise=2pt}, decorate] 
        (a) -- (b)
        node [pos=0.5, right=5pt, align=left, font=\tiny\bfseries, color=purple!80!black] {
            Faithfulness Score $s_2$ \\
            $\ell_{base} - \ell_{mod} = 3.87$
        };

\end{tikzpicture}
\caption{An example chain-of-thought intervention and score.}
\label{fig:vertical_pipeline}
\end{figure}

\subsection{Fine-Tuning}

\textbf{Data Generation.} For each question $q_j$ with ground truth
$a_{\text{true}, j}$, we sample a trace $(\mathbf{r}_j, a_j) \sim
    \mathcal{M}(\cdot \mid q_j)$ (temperatures listed in \cref{sec:experiments}), then
intervene on each step, generating modified traces $\{(\mathbf{r}_j^{(i)},
    a_j^{(i)})\}$ and scores (with $s_j^{(i)}$ corresponding to the trace with step
$i$ of $\mathbf{r}_j$ modified). We filter traces to be in the range
$[s_{\text{low}}, s_{\text{high}}]$ (low-scoring traces likely provide too weak
a signal to have a significant impact on faithfulness, and extremely
high-scoring traces are likely gibberish that decreases the log probability by
being nonsensical), then normalize to the range $[0.05, 1]$.

For $w_j^{(i)}$ greater than a certain threshold $\tau$, we append ground truth
answers to the trace for accuracy tuning.

Token $t$ receives weight
\begin{equation}
    \label{eqn:token-weights}
    \omega_j^{(i)}(t) = \begin{cases}
        0                    & \text{if input (prompt) token},                             \\
        0                    & \text{if reasoning token at or before $\tilde{r}_j^{(i)}$}, \\
        w_j^{(i)}            & \text{if reasoning token after $\tilde{r}_j^{(i)}$},        \\
        \eta \cdot w_j^{(i)} & \text{if answer token},
    \end{cases}
\end{equation}
where $\eta$ is a hyperparameter controlling how strongly to weight accuracy.

Assigning weights in this manner ensures the model isn't trained to generate
irrelevant facts, but rather to generate faithful reasoning \textit{after}
generating an incorrect step.

\textbf{Objective.} We optimize $\mathcal{L} = \mathcal{L}_{\text{faithfulness}} + \mathcal{L}_{\text{KL}}$, where
\begin{align}
    \mathcal{L}_{\text{faithfulness}} & = \frac{1}{|\mathcal{D}|} \sum_{(j,i) \in \mathcal{D}} \frac{\sum_t \omega_j^{(i)}(t) \ell_t}{\sum_t \omega_j^{(i)}(t)} \label{eqn:faith_loss}                                                          \\
    \mathcal{L}_{\text{KL}}           & = \frac{\beta}{|\mathcal{D}|} \sum_{(j,i) \in \mathcal{D}} \frac{\sum_t \omega_j^{(i)}(t) D_{\text{KL}}(p_{\mathcal{M}_{\text{ref}}} \| p_{\mathcal{M}})}{\sum_t \omega_j^{(i)}(t)} \label{eqn:kl_loss}
\end{align}
with $\ell_t$ the cross-entropy at token $t$, $\mathcal{M}_{\text{ref}}$ the
frozen base model (that is, the model before fine-tuning), and $\beta$
controlling regularization strength.

We fine-tune for one epoch using a LoRA adapter (necessary in order to fit in
VRAM). Hyperparameters (clustering $K$, LoRA rank, decoding temperatures, and
hyperparameters for fine-tuning) were chosen by hand-tuning to produce stable
training behavior on a small development set. The exact values used in all
experiments are listed in \cref{sec:experiments}.

\begin{algorithm}[h]
    \caption{LIFT Fine-Tuning}
    \label{alg:finetuning}
    \begin{algorithmic}[1]
        \REQUIRE Questions $\{q_j\}$, ground-truth answers $\{a_{\text{true}, j}\}$, model $\mathcal{M}$, weight $\beta$, score range $[s_{\text{low}}, s_{\text{high}}]$, answer score threshold $\tau$
        \ENSURE Fine-tuned model $\mathcal{M}'$
        \STATE $\mathcal{D} = \emptyset$
        \FOR{each $q_j$}
        \STATE $(\mathbf{r}_j, a_j) \sim \mathcal{M}(\cdot \mid q_j)$
        \STATE $\{(\mathbf{r}_j^{(i)}, a_j^{(i)}, s_j^{(i)})\} = \text{Score}(q_j, \mathbf{r}_j, a_j, \mathcal{M})$
        \FOR{$i$ with $s_j^{(i)} \in [s_{\text{low}}, s_{\text{high}}]$}
        \STATE Compute $w_j^{(i)}$ as $s_j^{(i)}$ normalized to range $[0.05, 1]$
        \IF{$w_j^{(i)} \geq \tau$}
        \STATE Append ``Answer: $a_{\text{true}, j}$'' to $\mathbf{r}_j^{(i)}$
        \ENDIF
        \STATE Compute token-level weights $\omega_j^{(i)}$ using \cref{eqn:token-weights}
        \STATE Append $(q_j, \mathbf{r}_j^{(i)}, \omega_j^{(i)})$ to $\mathcal{D}$
        \ENDFOR
        \ENDFOR
        \STATE Fine-tune $\mathcal{M}$ on $\mathcal{D}$ using \cref{eqn:faith_loss,eqn:kl_loss}
        \RETURN $\mathcal{M}'$
    \end{algorithmic}
\end{algorithm}

\section{Experimental Setup}
\label{sec:experiments}

We evaluate LIFT across multiple model architectures and reasoning benchmarks to
assess both faithfulness improvements and accuracy preservation. This section
describes our model selection, computational infrastructure, data generation
procedure, training configuration, evaluation methodology, and ablation study
design.

\textbf{Models.} We apply LIFT to three pretrained language models of varying
sizes and architectures: Mistral-7B-Instruct-v0.3, Qwen3-8B, and
Qwen3-4B-Instruct-2507. We apply LIFT using LoRA adapters~\citep{lora} with rank
$r=64$, LoRA alpha $\alpha=128$, and dropout $0.05$, targeting all linear
projection layers in the attention and feedforward blocks. This adapter-based
approach enables efficient fine-tuning while preserving the base model's general
capabilities.

\textbf{Training Data Generation.} We construct training datasets by sampling
from the train splits of six reasoning benchmarks: GSM8K \citep{gsm8k}, SVAMP
\citep{svamp}, StrategyQA \citep{strategyqa}, CommonsenseQA
\citep{commonsenseqa}, ASDiv \citep{asdiv}, and SciBench \citep{scibench}.
StrategyQA is under the CC BY 4.0 license, ASDiv is under a CC BY-NC 4.0
license, and the other four datasets are under the MIT license. All six datasets
are intended for model training and evaluation for CoT tasks. For each model, we
randomly select 1000 prompts from the combined pool of all six datasets. For
each selected question, we generate a preliminary reasoning trace using the
model. Temperature settings during this phase vary by model: we use temperature
$T=0.1$ for Mistral-7B-Instruct-v0.3 and $T=0.5$ for both Qwen models. These
settings balance the need for quality initial traces with sufficient diversity
across examples.

After generating preliminary traces, we apply Algorithm~\ref{alg:scoring} to
produce intervened traces and faithfulness scores for each reasoning step. When
generating continuations after intervened steps, we use higher temperatures to
encourage diverse reasoning paths: temperature $T=0.2$ for the Mistral model and
temperature $T=1.6$ for both Qwen models. The elevated temperature for
continuation generation increases the likelihood that the model will explore
alternative reasoning directions after the intervention rather than attempting
to recover the original answer despite the disrupted step.

We set $[s_{\text{low}}, s_{\text{high}}] = [0.5, 8]$, as traces with
faithfulness scores below $0.5$ typically ignore the intervened step or
self-correct, and those with scores above 8 are typically incoherent.

\textbf{Training Configuration.} We train using the AdamW optimizer with
learning rate $10^{-5}$ for all models. We use a per-device batch size of $4$
for the Mistral model and $2$ for both Qwen models, with gradient accumulation
over $2$ steps. All models are trained for one epoch using mixed precision
(fp16) to reduce memory consumption and accelerate training. The KL
regularization coefficient $\beta$ in Equation~\ref{eqn:kl_loss} is set to $0.8$
for Mistral-7B and $2.0$ for Qwen models.

We use different values of $\eta$ depending on the specific model and training
dataset (see \cref{tab:answer_weights}). We use one value for GSM8K, one for
StrategyQA and CommonsenseQA, and one for the remaining three datasets.

\begin{table}[h]
\centering
\small
\begin{tabular}{lccc}
\toprule
Model & GSM8K & QA Tasks & Other \\
\midrule
Mistral-7B & 1.8 & 0.4 & 0.8 \\
Qwen3-8B & 1.0 & 2.5 & 3.0 \\
Qwen3-4B & 1.0 & 2.5 & 3.0 \\
\bottomrule
\end{tabular}
\caption{$\eta$ hyperparameter by model and dataset type.}
\label{tab:answer_weights}
\end{table}

For each base model, we also fine-tune a \textit{baseline model}, with
final-answer-only examples through SFT. We use the same learning rate $10^{-5}$
and select $500$ prompts at random from each training dataset ($3000$ examples
in total). These models are used during evaluation to gauge the relative
accuracy and faithfulness of the LIFT-aligned models.

\begin{table*}[ht]
    \centering
    \caption{Performance comparison across all tested models. $\pm 2SE$ confidence intervals are reported. Model names are shortened for presentation. ``Trunc Faithfulness'' represents truncation faithfulness.}
    \label{tab:full_model_performance}
    \setlength{\tabcolsep}{12pt}
    \renewcommand{\arraystretch}{1.1}
    \small
    \begin{tabular}{lllccc}
        \toprule
        \textbf{Model} & \textbf{Method} & \textbf{Metric} & \textbf{GSM8K} & \textbf{SVAMP} & \textbf{StrategyQA} \\
        \midrule
        \multirow{7}{*}[-0.75em]{Mistral-7B} & \multirow{3}{*}{Aligned (CoT)} & Accuracy & $0.41 \pm 0.02$ & $0.69 \pm 0.02$ & $\mathbf{0.84 \pm 0.02}$ \\
         &  & Faithfulness & $\mathbf{1.71 \pm 0.21}$ & $\mathbf{2.01 \pm 0.28}$ & $\mathbf{0.72 \pm 0.23}$ \\
         &  & Static Faithfulness & $\mathbf{1.63 \pm 0.23}$ & $\mathbf{2.82 \pm 0.31}$ & $\mathbf{0.29 \pm 0.13}$ \\
         &  & Trunc Faithfulness & $\mathbf{5.32 \pm 0.52}$ & $\mathbf{4.55 \pm 0.70}$ & $\mathbf{0.64 \pm 0.14}$ \\
        \cmidrule{2-6}
         & \multirow{3}{*}{Baseline (CoT)} & Accuracy & $\mathbf{0.50 \pm 0.05}$ & $\mathbf{0.70 \pm 0.04}$ & $0.77 \pm 0.04$ \\
         &  & Faithfulness & $1.16 \pm 0.21$ & $1.04 \pm 0.18$ & $0.07 \pm 0.04$ \\
         &  & Static Faithfulness & $1.10 \pm 0.18$ & $2.05 \pm 0.36$ & $0.03 \pm 0.01$ \\
         &  & Trunc Faithfulness & $3.95 \pm 0.48$ & $2.91 \pm 0.64$ & $0.21 \pm 0.10$ \\
        \cmidrule{2-6}
        & Raw & Accuracy & $0.08 \pm 0.01$ & $0.49 \pm 0.03$ & $0.72 \pm 0.02$ \\
        \midrule
        \multirow{7}{*}[-0.75em]{Qwen3-8B} & \multirow{3}{*}{Aligned (CoT)} & Accuracy & $0.87 \pm 0.02$ & $0.95 \pm 0.01$ & $\mathbf{0.91 \pm 0.01}$ \\
         &  & Faithfulness & $\mathbf{1.33 \pm 0.16}$ & $\mathbf{0.98 \pm 0.25}$ & $\mathbf{0.45 \pm 0.09}$ \\
         &  & Static Faithfulness & $\mathbf{0.75 \pm 0.13}$ & $\mathbf{0.89 \pm 0.14}$ & $\mathbf{0.14 \pm 0.02}$ \\
         &  & Trunc Faithfulness & $\mathbf{4.66 \pm 0.49}$ & $\mathbf{3.95 \pm 0.52}$ & $\mathbf{0.35 \pm 0.07}$ \\
        \cmidrule{2-6}
         & \multirow{3}{*}{Baseline (CoT)} & Accuracy & $\mathbf{0.90 \pm 0.03}$ & $\mathbf{0.97 \pm 0.02}$ & $0.87 \pm 0.03$ \\
         &  & Faithfulness & $0.58 \pm 0.10$ & $0.54 \pm 0.20$ & $0.09 \pm 0.03$ \\
         &  & Static Faithfulness & $0.43 \pm 0.12$ & $0.36 \pm 0.12$ & $0.07 \pm 0.03$ \\
         &  & Trunc Faithfulness & $3.47 \pm 0.44$ & $1.93 \pm 0.44$ & $0.15 \pm 0.04$ \\
        \cmidrule{2-6}
        & Raw & Accuracy & $0.17 \pm 0.02$ & $0.68 \pm 0.02$ & $0.40 \pm 0.03$ \\
        \midrule
        \multirow{7}{*}[-0.75em]{Qwen3-4B} & \multirow{3}{*}{Aligned (CoT)} & Accuracy & $\mathbf{0.90 \pm 0.01}$ & $\mathbf{0.93 \pm 0.01}$ & $0.86 \pm 0.02$ \\
         &  & Faithfulness & $\mathbf{1.12 \pm 0.18}$ & $\mathbf{0.91 \pm 0.23}$ & $\mathbf{0.58 \pm 0.09}$ \\
         &  & Static Faithfulness & $\mathbf{0.71 \pm 0.13}$ & $\mathbf{0.83 \pm 0.17}$ & $\mathbf{0.12 \pm 0.03}$ \\
         &  & Trunc Faithfulness & $\mathbf{3.87 \pm 0.43}$ & $\mathbf{3.16 \pm 0.50}$ & $\mathbf{0.48 \pm 0.10}$ \\
        \cmidrule{2-6}
         & \multirow{3}{*}{Baseline (CoT)} & Accuracy & $\mathbf{0.90 \pm 0.03}$ & $\mathbf{0.93 \pm 0.02}$ & $\mathbf{0.87 \pm 0.03}$ \\
         &  & Faithfulness & $0.59 \pm 0.10$ & $0.47 \pm 0.16$ & $0.03 \pm 0.01$ \\
         &  & Static Faithfulness & $0.30 \pm 0.06$ & $0.37 \pm 0.12$ & $0.05 \pm 0.03$ \\
         &  & Trunc Faithfulness & $2.54 \pm 0.37$ & $1.54 \pm 0.45$ & $0.15 \pm 0.08$ \\
        \cmidrule{2-6}
        & Raw & Accuracy & $0.33 \pm 0.02$ & $0.70 \pm 0.02$ & $0.57 \pm 0.02$ \\
        \bottomrule
    \end{tabular}
\end{table*}

\begin{figure*}[ht]
    \begin{subfigure}{\textwidth}
        \includegraphics[width=\textwidth]{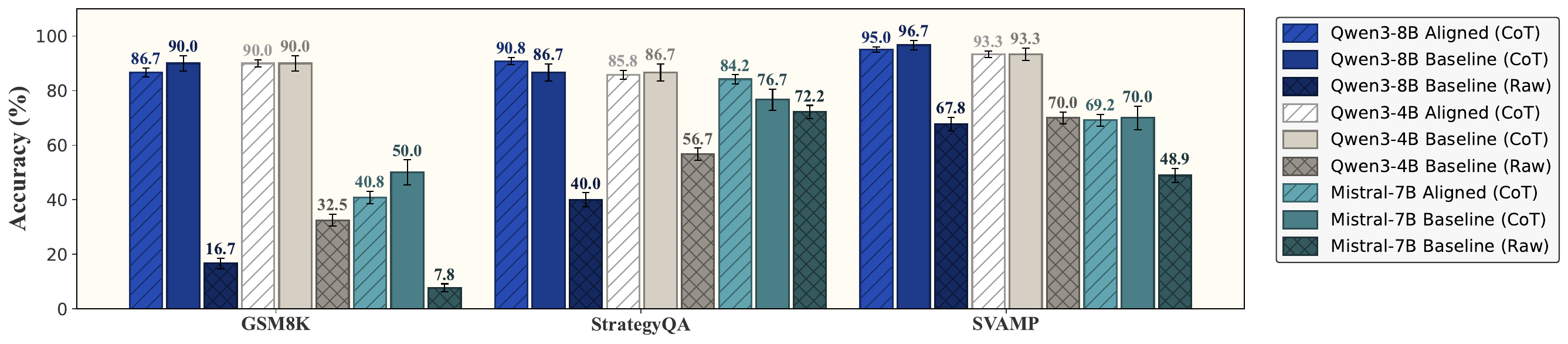}
    \end{subfigure}
    \begin{subfigure}{\textwidth}
        \includegraphics[width=\textwidth]{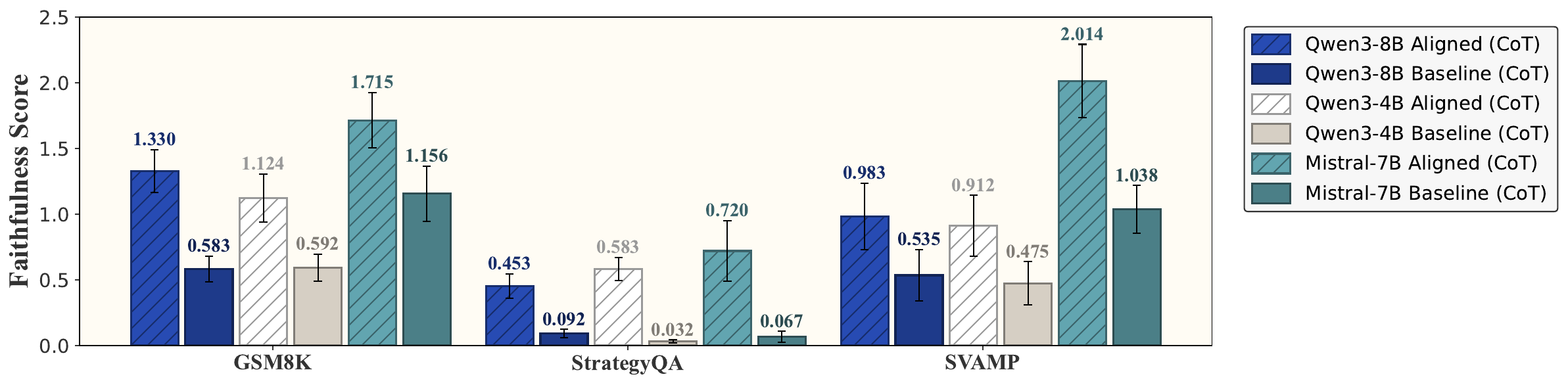}
    \end{subfigure}
    \caption{Accuracy and faithfulness across all models and datasets.
        ``Baseline'' represents the baseline model fine-tuned solely for
        accuracy, ``Aligned'' represents the model after fine-tuning using LIFT.
        ``CoT'' represents chain-of-thought prompting, and ``Raw'' represents
        the baseline model with no chain-of-thought.}
    \label{fig:acc-faith}
\end{figure*}

\begin{table*}[t]
    \centering
    \caption{Ablation of answer weight $\lambda$ across all models. Faithfulness
    is the mean faithfulness score across all tested traces from that dataset;
    Accuracy is the mean exact-match accuracy.}
    \label{tab:ablation_results}
    \small
    \setlength{\tabcolsep}{8pt}
    \begin{tabular}{ll ccc ccc}
        \toprule
                                    &           & \multicolumn{2}{c}{GSM8K} & \multicolumn{2}{c}{StrategyQA} & \multicolumn{2}{c}{SVAMP}                          \\
        \cmidrule(lr){3-4} \cmidrule(lr){5-6} \cmidrule(lr){7-8}
        Model                       & $\lambda$ & Faithfulness              & Accuracy                       & Faithfulness                  & Accuracy  & Faithfulness & Accuracy  \\
        \midrule
        \multirow{3}{*}{Mistral-7B} & 0.0 & 2.291 & 0.241 & 0.831 & 0.658 & 2.687 & 0.567 \\
                                    & 0.5       & 1.862                     & 0.433                          & 0.782                         & 0.767     & 2.228        & 0.633     \\
                                    & 1.0       & 1.715                     & 0.408                          & 0.720                         & 0.842     & 2.014        & 0.692     \\
        \midrule
        \multirow{3}{*}{Qwen3-8B}   & 0.0       & 3.138                     & 0.392                          & 1.019                         & 0.600     & 2.023        & 0.650     \\
                                    & 0.5       & 1.809                     & 0.817                          & 0.447                         & 0.900     & 1.032        & 0.958     \\
                                    & 1.0       & 1.330                     & 0.867                          & 0.453                         & 0.908     & 0.983        & 0.950     \\
        \midrule
        \multirow{3}{*}{Qwen3-4B}   & 0.0 & 1.398 & 0.767 & 0.661 & 0.833 & 1.239 & 0.867 \\
                                    & 0.5       & 1.356                     & 0.858                          & 0.537                         & 0.900     & 1.021        & 0.867     \\
                                    & 1.0       & 1.124                     & 0.900                          & 0.583                         & 0.858     & 0.912        & 0.933     \\
        \bottomrule
    \end{tabular}
\end{table*}

\textbf{Evaluation Benchmarks and Protocol.} We evaluate models on the test
splits of GSM8K, SVAMP, and StrategyQA. These benchmarks span different
reasoning types and difficulty levels. GSM8K and SVAMP require multi-step
arithmetic with numerical answers, while StrategyQA requires fact composition
with boolean answers. We use the first $480$ entries from the test splits of
GSM8K and StrategyQA, and the full $300$ entries from the test split of SVAMP.

For each model, we evaluate three configurations: the baseline model prompted
with chain-of-thought reasoning, the baseline model prompted without
chain-of-thought (direct answer prediction), and the LIFT-aligned model prompted
with chain-of-thought reasoning. All evaluations use greedy decoding to ensure
deterministic outputs. We assess answer accuracy by comparing the model's final
answer to the ground truth using exact-match comparison. For GSM8K and SVAMP, we
extract the final numeric value from the model's output. For StrategyQA, we
extract the boolean answer (True or False) via case-insensitive pattern
matching.

\textbf{Faithfulness Measurement.} For all chain-of-thought evaluations, we
measure faithfulness using our intervention-based metric from
Algorithm~\ref{alg:scoring}. For each reasoning step $r_i$ in a generated trace
$\mathbf{r} = (r_1, \ldots, r_n)$, we apply Algorithm~\ref{alg:intervention} to
produce an intervened step $\tilde{r}_i$. For each intervened step, we generate
a continuation and compute the faithfulness score $s_i$ as specified in
Equation~\ref{eq:faithfulness_score}. The overall faithfulness score for a trace
is computed as the mean of stepwise scores.

In addition to our primary faithfulness metric, we evaluate static faithfulness.
Static faithfulness measures the log-probability drop when replacing step $r_i$
with $\tilde{r}_i$ but keeping all subsequent steps $r_{i+1}, \ldots, r_n$
unchanged from the original trace. Static faithfulness provides a more
conservative estimate of causal influence, measuring whether a step's specific
content directly affects the answer probability even when forcing the model to
continue with the original reasoning path. This contrasts with our primary
metric, which allows the model to naturally diverge after intervention. We
provide additional discussion of static faithfulness and its interpretation in
Appendix~\ref{app:static_faith}.

We also evaluate truncation faithfulness. Truncation faithfulness measures the
log-probability drop when removing all steps from $r_i$ onward and preserving
all preceding steps. This faithfulness measure, described in
\citet{measuringfaithfulness}, provides a faithfulness measure that does not
rely on counterfactual edits, avoiding any potential artifacts caused by
sensitivity to counterfactual edits. Truncation faithfulness is discussed in
more detail in Appendix~\ref{app:trunc_faith}.

We provide details on computational cost and approximate times in
Appendix~\ref{app:repr}.

\section{Results}

Our evaluation results are summarized in \cref{tab:full_model_performance}.
\cref{fig:acc-faith} displays accuracy and faithfulness results. The
LIFT-aligned models show increases in faithfulness compared to the base model
across all base models and datasets, statistically significant across all model
and benchmark combinations tested (one-tailed paired $t$-test, $p < 0.05$).

These results demonstrate a marked improvement in faithfulness, with relative
increases in faithfulness ranging from 48\% to 1722\% depending on the model and
benchmark. All models show large gains on StrategyQA, ranging from 392\% to
1722\% relative increases in faithfulness. On GSM8K and SVAMP, all models show
roughly a 100\% increase in faithfulness.

Critically, these faithfulness improvements are achieved without compromising
task performance. Across all three models, we find no significant change in
accuracy between the base model with CoT and the aligned model with CoT.

As defined in Appendix~\ref{app:static_faith}, static faithfulness measures the
log-probability drop of the answer when a step is replaced with a counterfactual
edit while keeping the subsequent steps fixed. The results in
\Cref{tab:full_model_performance} show a consistent increase in static
faithfulness across all evaluated models and benchmarks. In Mistral-7B, static
faithfulness improved from 0.03 to 0.29 on StrategyQA and from 2.05 to 2.82 on
SVAMP. Similarly, Qwen3-4B demonstrated a gain from 0.05 to 0.12 on StrategyQA.
An increase in this metric shows that after LIFT fine-tuning, individual steps
have a greater impact on the final answer, even without changing the steps that
follow.

As explained in Appendix~\ref{app:trunc_faith}, truncation faithfulness measures
the log-probability drop of the answer when a step and all subsequent steps are
removed, leaving preceding steps unchanged. The results in
\Cref{tab:full_model_performance} show consistent and significant gains in
truncation faithfulness on all models and benchmarks.

These improvements are noteworthy because LIFT does not explicitly train for
static faithfulness or truncation faithfulness. This addresses the risk of
circular evaluation: As described in \cref{alg:finetuning}, the fine-tuning data
$\mathcal{D}$ consists of modified traces where the model generates a new
continuation following an intervention. Consequently, the model is trained to
maintain logical consistency during active regeneration rather than under the
fixed-suffix conditions used in static evaluation or the early-stopping
conditions in truncation faithfulness. The observed gains in these two measures
suggest that the faithfulness gains through LIFT lead to generalization beyond
one specific faithfulness measure. In particular, the gains in truncation
faithfulness indicate that the faithfulness gains are not limited to
intervention-based measures.

\subsection{Ablation Study}

To characterize the trade-off between faithfulness and accuracy in LIFT, we
perform an ablation study over a parameter $\lambda$, which scales the
coefficient $\eta$ in Equation~\ref{eqn:token-weights}. We train each model with
three configurations: $\lambda = 0.0$ (pure faithfulness training with no answer
supervision), $\lambda = 1.0$ (balanced training using our default answer weight
settings; this is the same as the aligned models shown in
\cref{tab:full_model_performance}), and $\lambda = 0.5$, weighing answers half
as strongly as the balanced training. The $\lambda = 0.0$ configuration shows
the level of faithfulness that can be achieved using LIFT if accuracy
degradation is not a concern, while $\lambda = 1.0$ is our recommended default
that balances faithfulness improvements with answer accuracy preservation.

For each ablation configuration, we follow the same data generation and training
procedures described in \cref{sec:method}, changing only the hyperparameter
$\eta$ to $\lambda \cdot \eta$. We evaluate all configurations on the same test
sets using identical faithfulness and accuracy metrics.

The ablation results are shown in \cref{tab:ablation_results}. They provide
empirical evidence for a controllable trade-off between the faithfulness of
reasoning steps and the model's final answer accuracy. At $\lambda = 0.0$, where
the model is fine-tuned solely on faithfulness, we observe the highest
faithfulness scores across all datasets. For example, Mistral-7B achieves a
faithfulness score of 2.291 on GSM8K at $\lambda = 0.0$, compared to 1.715 at
$\lambda = 1.0$. However, this focus on faithfulness significantly reduces
accuracy, particularly on arithmetic tasks like GSM8K, where Mistral's accuracy
drops to 0.241.

As $\lambda$ increases, the inclusion of ground-truth answer supervision allows
the models to recover significant task performance. In the case of Qwen3-8B on
GSM8K, moving from $\lambda = 0.0$ to $\lambda = 0.5$ results in a 47.5-point
increase in accuracy (0.392 to 0.867), albeit with a drop in faithfulness, from
3.138 to 1.330. This suggests that a greater $\lambda$ results in greater
accuracy at the cost of lower faithfulness.

This ablation demonstrates that LIFT provides a mechanism to adjust the
faithfulness-accuracy trade-off based on application requirements. Applications
requiring maximum interpretability and transparency may prefer higher
faithfulness at some accuracy cost, while applications prioritizing task
performance may prefer configurations that maintain stronger answer supervision.

\subsection{Qualitative Example}

In order to illustrate the differences in reasoning style between models aligned
with LIFT and base models, we provide an example of reasoning traces from
Qwen3-4B-Instruct-2507 responding to a chain-of-thought prompt. Faithfulness
scores per step are shown.

\hspace{-1.2em}
\fbox{\begin{minipage}{\linewidth}
\small
\textbf{Question:} Happy hardcore is a music genre of hard dance. Happy hardcore
emerged both from the UK breakbeat hardcore rave scene, and Belgian, German and
Dutch hardcore techno scenes. A funeral is traditionally a somber event.
Funerals typically do not involve dancing. Raves are typically energetic and
upbeat places and are not somber like a funeral. Would it be unusual to play
Happy hardcore music at a funeral?
\end{minipage}}
\begin{minipage}[t]{\linewidth}
\fbox{\begin{minipage}{\textwidth}
\small
\textbf{Base Model (Qwen3-4B-Instruct-2507)} \\[0.2cm]
\textit{Reasoning:}
\begin{enumerate}
    \item Happy hardcore is a music genre that is energetic and upbeat. \textcolor{blue!70!black}{[$s_1 = 0.000$]}
    \item Funerals are traditionally somber events. \textcolor{blue!70!black}{[$s_2 = 0.000$]}
    \item The genre is not typically associated with funerals. \textcolor{blue!70!black}{[$s_3 = 0.000$]}
    \item Therefore, it is unusual to play Happy hardcore music at a funeral. \textcolor{blue!70!black}{[$s_4 = 0.000$]}
\end{enumerate}
\textbf{Answer:} True \\[0.1cm]
\textit{Mean: 0.000}
\end{minipage}}
\end{minipage}%
\hfill
\begin{minipage}[t]{\linewidth}
\fbox{\begin{minipage}{\textwidth}
\small
\textbf{Aligned Model (Qwen3-4B-Instruct-2507, LIFT-aligned)} \\[0.2cm]
\textit{Reasoning:}
\begin{enumerate}
    \item Happy hardcore is a music genre that is associated with energetic, upbeat, and dance-oriented events such as raves. \textcolor{blue!70!black}{[$s_1 = 2.976$]}
    \item Funerals are traditionally somber and not associated with dancing. \textcolor{blue!70!black}{[$s_2 = 3.058$]}
    \item Therefore, playing Happy hardcore music at a funeral would be unusual. \textcolor{blue!70!black}{[$s_3 = 3.016$]}
\end{enumerate}
\textbf{Answer:} True \\[0.1cm]
\textit{Mean: 3.017}
\end{minipage}}
\end{minipage}

This example illustrates the difference in reasoning behavior between the two
models. While both models produce correct answers, their reasoning traces have
different causal structures. The base model generates four reasoning steps with
faithfulness scores of zero (up to 11 digits of precision)---intervening on any
step does not change the model's final answer probability. This suggests that
the answer is predetermined, with the reasoning serving as a post-hoc
explanation rather than causally determining the conclusion.

In contrast, the LIFT-aligned model produces a more concise three-step trace
where each step exhibits measurable causal influence (mean faithfulness 3.017).
Beyond the quantitative improvement, the aligned model demonstrates qualitative
differences in reasoning structure. Each step explicitly builds on preceding
information: the first step establishes the genre's association with
dance-oriented events, the second introduces the incompatibility with funeral
contexts by noting that funerals are ``not associated with dancing,'' and the
third synthesizes these facts into the final conclusion. This sequential
dependency structure reflects the causal necessity that LIFT optimizes for
during training.

\section{Future Work}
\label{sec:future}

Several promising directions extend LIFT's capabilities.

\textbf{Multi-Step Interventions.} Current scoring intervenes on single steps
independently. Future work could explore joint interventions on multiple steps
simultaneously, measuring whether step combinations are collectively necessary.
This could reveal redundancy in reasoning traces and identify minimal sufficient
subsets.

\textbf{Adaptive Intervention Strategies.} Rather than using a fixed fact
corpus, future methods could generate counterfactual edits on-the-fly using
language models, conditioning on the specific reasoning context. This could
improve intervention quality in specialized domains and adapt to novel reasoning
patterns.

\textbf{Scaling to Multi-Modal Reasoning.} As language models increasingly
incorporate vision and other modalities, extending LIFT's intervention-based
approach to multi-modal reasoning traces presents an important challenge. This
would require developing appropriate counterfactual interventions for visual
evidence and other non-textual reasoning components.

\section{Software and Data}

For reproducibility we provide the complete list of hyperparameters, prompt
templates, the fact corpus and cluster assignments, and data files containing
the examples used for fine-tuning the models. We provide code to generate
examples using LIFT and fine-tune for faithfulness, or fine-tune directly on the
examples that we used. All files can be found in the supplementary material.

\newpage
\section{Limitations}
\label{sec:limitations}

While LIFT demonstrates consistent faithfulness improvements, several
limitations warrant discussion.

\textbf{Computational Cost.} LIFT requires generating multiple modified traces
per training example (one per reasoning step), making data generation
significantly more expensive than standard supervised fine-tuning. For a trace
with $n$ steps, we require $n$ forward passes for scoring plus $n$ generation
passes for continuations.

\textbf{Base Model Capability Requirement.} LIFT assumes the base model can
already generate coherent chain-of-thought reasoning. For models that produce
incoherent or very short traces, the intervention-based scoring may not provide
meaningful signal. Future work could explore bootstrapping approaches for models
with weaker initial reasoning capabilities.

\textbf{Fact Corpus Dependency.} Our intervention quality depends on the fact
corpus $\mathcal{F}$ containing domain-appropriate facts. For highly specialized
domains (e.g.\ advanced mathematics, medical reasoning), the corpus may lack
suitable counterfactuals, potentially reducing intervention effectiveness.
Future work could explore alternative methods of retrieving facts for
interventions.

\textbf{Hyperparameter Selection.} Due to computational constraints,
hyperparameters were selected through manual tuning on a small development set
rather than systematic grid search. The relationships between hyperparameters
(temperature settings, score thresholds, LoRA rank) and final performance remain
underexplored. Future work could investigate these dependencies more rigorously
to optimize the faithfulness-accuracy trade-off and identify which
hyperparameters are most critical for different model families or task types.

\textbf{Metric Appropriateness.} The metrics we use (faithfulness, static
faithfulness, truncation faithfulness) are proxies for faithfulness and may not
capture it fully, an unavoidable consequence of avoiding human annotation. The
consistency of improvements across all three metrics, which differ in their
intervention mechanism, provides some evidence that the gains reflect a true
underlying change in reasoning behavior rather than overfitting to a single
measurement approach.

\textbf{Risk.} Because LIFT optimizes proxy measures of faithfulness rather than
direct human judgments, improvements in score may not always correspond to
genuinely faithful reasoning. As a result, a model fine-tuned with LIFT could
appear more faithful than it is. In safety-critical settings, LIFT should
therefore be used only with additional human review or independent faithfulness
validation.

\bibliography{main.bib}

@article{wei2022cot,
  title={Chain-of-thought prompting elicits reasoning in large language models},
  author={Wei, Jason and Wang, Xuezhi and Schuurmans, Dale and Bosma, Maarten and Xia, Fei and Chi, Ed and Le, Quoc V and Zhou, Denny and others},
  journal={Advances in neural information processing systems},
  volume={35},
  pages={24824--24837},
  year={2022}
}

@inproceedings{lightman2023lets,
  title={Let's verify step by step},
  author={Lightman, Hunter and Kosaraju, Vineet and Burda, Yuri and Edwards, Harrison and Baker, Bowen and Lee, Teddy and Leike, Jan and Schulman, John and Sutskever, Ilya and Cobbe, Karl},
  booktitle={The Twelfth International Conference on Learning Representations},
  year={2023}
}

@misc{cotnotexplain,
  author = {Barez, Fazl and Wu, Tung-Yu and Arcuschin, Iván and Lan, Michael and Wang, Vincent and Siegel, Noah and Collignon, Nicolas and Neo, Clement and Lee, Isabelle and Paren, Alasdair and Bibi, Adel and Trager, Robert and Fornasiere, Damiano and Yan, John and Elazar, Yanai and Bengio, Yoshua},
  title = {Chain-of-Thought Is Not Explainability},
  publisher = {alphaXiv},
  year = {2025}
}

@inproceedings{frodo2024,
  title={Making Reasoning Matter: Measuring and Improving Faithfulness of Chain-of-Thought Reasoning},
  author={Paul, Debjit and West, Robert and Bosselut, Antoine and Faltings, Boi},
  booktitle={Findings of the Association for Computational Linguistics: EMNLP 2024},
  pages={15012--15032},
  year={2024}
}

@article{measuringfaithfulness,
  title={Measuring Faithfulness in Chain-of-Thought Reasoning},
  author={Lanham, Tamera and Chen, Anna and Radhakrishnan, Ansh and Steiner, Benoit and Denison, Carson and Hernandez, Danny and Li, Dustin and Durmus, Esin and Hubinger, Evan and Kernion, Jackson and others},
  journal={CoRR},
  year={2023}
}

@article{gsm8k,
  title={Training verifiers to solve math word problems},
  author={Cobbe, Karl and Kosaraju, Vineet and Bavarian, Mohammad and Chen, Mark and Jun, Heewoo and Kaiser, Lukasz and Plappert, Matthias and Tworek, Jerry and Hilton, Jacob and Nakano, Reiichiro and others},
  journal={arXiv preprint arXiv:2110.14168},
  year={2021}
}

@inproceedings{svamp,
  title={Are NLP Models really able to Solve Simple Math Word Problems?},
  author={Patel, Arkil and Bhattamishra, Satwik and Goyal, Navin},
  booktitle={Proceedings of the 2021 Conference of the North American Chapter of the Association for Computational Linguistics: Human Language Technologies},
  pages={2080--2094},
  year={2021}
}

@article{strategyqa,
  title={Did aristotle use a laptop? a question answering benchmark with implicit reasoning strategies},
  author={Geva, Mor and Khashabi, Daniel and Segal, Elad and Khot, Tushar and Roth, Dan and Berant, Jonathan},
  journal={Transactions of the Association for Computational Linguistics},
  volume={9},
  pages={346--361},
  year={2021},
  publisher={MIT Press One Rogers Street, Cambridge, MA 02142-1209, USA journals-info~…}
}

@inproceedings{commonsenseqa,
  title={Commonsenseqa: A question answering challenge targeting commonsense knowledge},
  author={Talmor, Alon and Herzig, Jonathan and Lourie, Nicholas and Berant, Jonathan},
  booktitle={Proceedings of the 2019 Conference of the North American Chapter of the Association for Computational Linguistics: Human Language Technologies, Volume 1 (Long and Short Papers)},
  pages={4149--4158},
  year={2019}
}

@inproceedings{asdiv,
  title={A diverse corpus for evaluating and developing English math word problem solvers},
  author={Miao, Shen-Yun and Liang, Chao-Chun and Su, Keh-Yih},
  booktitle={Proceedings of the 58th annual meeting of the Association for Computational Linguistics},
  pages={975--984},
  year={2020}
}

@inproceedings{scibench,
  title={SciBench: Evaluating College-Level Scientific Problem-Solving Abilities of Large Language Models},
  author={Wang, Xiaoxuan and Hu, Ziniu and Lu, Pan and Zhu, Yanqiao and Zhang, Jieyu and Subramaniam, Satyen and Loomba, Arjun R and Zhang, Shichang and Sun, Yizhou and Wang, Wei},
  booktitle={International Conference on Machine Learning},
  pages={50622--50649},
  year={2024},
  organization={PMLR}
}

@inproceedings{lora,
  title={LoRA: Low-Rank Adaptation of Large Language Models},
  author={Hu, Edward J and Wallis, Phillip and Allen-Zhu, Zeyuan and Li, Yuanzhi and Wang, Shean and Wang, Lu and Chen, Weizhu and others},
  booktitle={International Conference on Learning Representations},
  year={2022}
}

@article{unfaithful,
  title={Language models don't always say what they think: Unfaithful explanations in chain-of-thought prompting},
  author={Turpin, Miles and Michael, Julian and Perez, Ethan and Bowman, Samuel},
  journal={Advances in Neural Information Processing Systems},
  volume={36},
  pages={74952--74965},
  year={2023}
}

@inproceedings{faithfulcot,
  title={Faithful chain-of-thought reasoning},
  author={Lyu, Qing and Havaldar, Shreya and Stein, Adam and Zhang, Li and Rao, Delip and Wong, Eric and Apidianaki, Marianna and Callison-Burch, Chris},
  booktitle={Proceedings of the 13th International Joint Conference on Natural Language Processing and the 3rd Conference of the Asia-Pacific Chapter of the Association for Computational Linguistics (Volume 1: Long Papers)},
  pages={305--329},
  year={2023}
}

\newpage
\newpage
\clearpage
\appendix

\section{Fact Corpus Construction and Clustering}
\label{app:clustering}

\textbf{Corpus Sources.} Our fact corpus $\mathcal{F}$ contains approximately 350,000 entries from two sources:
\begin{itemize}
    \item \textbf{Wikidata facts} (250,000 entries): We extract factual triples from Wikidata and convert them to natural language using a language model. Examples include geographic facts (``Tokyo is the capital of Japan''), temporal facts (``World War II ended in 1945''), and relational facts (``Albert Einstein developed the theory of relativity'').
    \item \textbf{Arithmetic equations} (100,000 entries): We generate simple arithmetic expressions covering addition, subtraction, multiplication, and division with operands ranging from $-1000$ to 1000. Each equation is formatted in multiple styles: symbolic (``143 - 61 = 82''), natural language (``105 + 106 equals 211.''), and mixed formats.
\end{itemize}

\textbf{Embedding and Clustering.} We compute sentence embeddings using the
BGE-large-en-v1.5 model, which produces 1024-dimensional vectors. We apply
k-means clustering with 20 initial clusters and $K=5000$ target clusters, chosen
to balance granularity (enabling domain-specific matching) and coverage
(ensuring sufficient facts per cluster).

The same fact corpus was used for interventions for all models and runs. All
facts and clusters are provided in the codebase.

\section{Style Rewriting Implementation}
\label{app:style_rewrite}

To maintain coherence with the original CoT context, we rewrite the retrieved counterfactual fact $f^*$ to match the writing style of the original step $r_i$. We use the base model itself to perform this rewriting via few-shot prompting.

\textbf{Prompt Structure.} The prompt contains 16 few-shot examples demonstrating various style transformations:
\begin{itemize}
    \item Formal to informal: ``Seven plus two equals nine'' $\rightarrow$ ``In math class today, we discovered that seven plus two makes nine''
    \item Symbolic to code: ``6 × 7 = 42'' $\rightarrow$ ``\# compute product\textbackslash nresult = 6 * 7''
    \item Natural language to symbolic: ``Nine minus four equals five'' $\rightarrow$ ``9 - 4 = 5''
    \item LaTeX formatting: ``8 / 2 = 4'' $\rightarrow$ ``\$8 \textbackslash div 2 = 4\$''
\end{itemize}

The prompt explicitly instructs the model to rewrite the step such that its writing style flows with the original step but the content contradicts it. This ensures $\tilde{r}_i$ is contextually coherent but factually incorrect.

\textbf{Generation Parameters.} We use greedy decoding for deterministic style rewriting. The model generates a chain-of-thought explanation, followed by the final rewritten step on a separate line, which we extract via pattern matching.

\newpage
\onecolumn

\section{Static Faithfulness Metric}
\label{app:static_faith}

In addition to our main faithfulness metric (which regenerates continuations), we evaluate \emph{static faithfulness}: the log-probability drop when replacing a step but keeping all subsequent steps unchanged.

\textbf{Definition.} For step $i$ in trace $\mathbf{r} = (r_1, \ldots, r_n)$, static faithfulness is:
\begin{equation}
    s_i^{\text{static}} = \log p(a \mid q, \mathbf{r}) - \log p(a \mid q, r_1, \ldots, r_{i-1}, \tilde{r}_i, r_{i+1}, \ldots, r_n)
\end{equation}
where $\tilde{r}_i$ is the intervened step from Algorithm \ref{alg:intervention}.

\textbf{Interpretation.} Static faithfulness measures whether a step's specific content affects the answer, even when forcing the same continuation. High static faithfulness indicates the model uses the step's information to support the answer. Low static faithfulness suggests the step is decorative—changing its content doesn't affect answer probability because subsequent steps override it.

\textbf{Comparison.} Our main (regenerative) faithfulness metric typically
yields higher scores because it allows the model to naturally diverge after
intervention. Static faithfulness is more conservative, only measuring direct
influence on the answer. In particular, if the model's answer is calculated in
the final step of the CoT, there is little to no drop in log probability of the
answer if a previous step is modified. Nevertheless, we report both metrics in
our evaluation to provide complementary views of causal importance.

\section{Truncation Faithfulness Metric}
\label{app:trunc_faith}

The final faithfulness metric that we measure is \emph{truncation faithfulness}:
the log-probability drop when removing a step and all following steps. This
faithfulness measure is proposed in \citet{measuringfaithfulness}.

\textbf{Definition.} For step $i$ in trace $\mathbf{r} = (r_1, \ldots, r_n)$, truncation faithfulness is:
\begin{equation}
    s_i^{\text{trunc}} = \log p(a \mid q, \mathbf{r}) - \log p(a \mid q, r_1, \ldots, r_{i-1}).
\end{equation}

\textbf{Interpretation.} Truncation faithfulness measures the cumulative causal
importance of step $i$ and all subsequent steps: it asks how much the model's
answer depends on having reasoning from step $i$ onward at all. High truncation
faithfulness for step $i$ indicates that the reasoning from step $i$ to the end
of the trace is collectively necessary to produce the correct answer. Low
truncation faithfulness suggests the answer is already well-determined by the
prefix $r_1, r_2, \dots, r_{i-1}$.

\textbf{Comparison.} Truncation faithfulness serves in our experiments as a
measure of faithfulness independent of our intervention procedure. The
truncation procedure mitigates any confounding that may be introduced by the
intervened steps (for example, distribution shift of following steps and the
final answer due to an unnatural or inconsistent step).

\newpage
\twocolumn

\section{Reproducibility Notes}
\label{app:repr}

Data generation is parallelized across multiple worker threads to improve
efficiency. For Mistral-7B-Instruct-v0.3, we use two worker threads on an NVIDIA
H100 SXM GPU (80GB VRAM). For Qwen3-8B, we use two worker threads on an NVIDIA
GH200 GPU (96GB VRAM). For Qwen3-4B-Instruct-2507, we use three worker threads
on an NVIDIA GH200 GPU.

Data generation takes approximately 24 hours (wall-clock time) for
Mistral-7B-Instruct-v0.3 and Qwen3-8B, and approximately 10 hours for
Qwen3-4B-Instruct-2507. Fine-tuning on the generated 1000 examples takes
approximately 10 minutes for Mistral-7B-Instruct-v0.3 and Qwen3-8B, and
approximately 6 minutes for Qwen3-4B-Instruct-2507. Total end-to-end training
time (including data generation and fine-tuning) is approximately 24 hours for
the 7-8B models and 10 hours for the 4B model on the specified hardware.

Evaluation of 300-480 examples per dataset with faithfulness scoring takes
approximately 4-5 hours per model per dataset. The total computational cost for
reproducing all experiments (3 models, 3 datasets) is approximately 100
GPU-hours on H100/GH200 class hardware.

\end{document}